\documentclass[journal]{IEEEtran}
\usepackage{times}
\usepackage{helvet}
\usepackage{courier}
\usepackage{subfigure}
\usepackage{url}
\usepackage{graphicx}
\usepackage{amsmath}
\usepackage{multirow}
\usepackage{algorithm}
\usepackage{algorithmic}
\usepackage{indentfirst}
\ifCLASSINFOpdf
\else
\fi
\begin{document}
%
\title{LSTM-based Flow Prediction}
%
%
%

\author{Hongzhi~Wang~\IEEEmembership{Member,~IEEE,}
        Yang~Song,
        and~Shihan~Tang
\thanks{wangzh@hit.edu.cn, Department of Computer Science, Harbin Institute of Technology, Harbin, Heilongjiang 150001, \underline{China}}}

%
%

\markboth{}%
{Shell \MakeLowercase{\textit{et al.}}: Bare Demo of IEEEtran.cls for IEEE Journals}
%



\maketitle

\begin{abstract}
In this paper, a method of prediction on continuous time series variables from the production or flow - an LSTM algorithm based on multivariate tuning -  is proposed. The algorithm improves the traditional LSTM algorithm and converts the time series data into supervised learning sequences regarding industrial data's features. The main innovation of this paper consists in introducing the concepts of periodic measurement and time window in the industrial prediction problem, especially considering industrial data with time series characteristics. Experiments using real-world datasets show that the prediction accuracy is improved, 54.05\% higher than that of traditional LSTM algorithm.
\end{abstract}

%
\IEEEpeerreviewmaketitle

\section{Introduction}\label{introduction}
\indent In industry, with the high-speed functioning of the enterprise product line, data are generated continuously. 
Malfunctions and abnormality often take place, which incur a great deal of money and resources, and even advanced equipment cannot avoid these problems\cite{jardine2006review}. Industrial companies have to pay a lot to maintain and ensure the normal operation of the manufacturing process. According to the statistics, the maintenance costs of all kinds of industrial enterprises account for about 15\%-70\% of total production costs\cite{bevilacqua2000analytic}.\\
\indent Flow prediction is motivated by such industrial conundrums faced by many factories. Implementing flow prediction to forecast the output of the machine and to detect the problems in time via prediction, not only can production increase, but also a large number of workforce and resources for troubleshooting can be saved. Thus, engineers and scholars have focused their attention on how to detect the failure of the manufacturing process and predict the production or flow to ensure the well-functioning of the equipment.\\
\indent However, challenges are met, especially that industrial time series data diverge from conventional ones. First, these data are most collected by sensors with high sampling rate, so pieces of information is obtained with in a second, namely these data have short intervals. Therefore, the problem of long-term dependencies is more likely to be exposed and not all deep learning methods are able to solve it. Besides, various types of sensors are used at the same time, by which the data collected can be correlated rather than independent. For example, pressure depends on temperature in hermetic space. Thus, many dependent dimensions should be considered. Moreover, industrial data also have an opaque periodicity so that such a characteristic is hard to detect and usually neglected, which diminishes the accuracy of predictive algorithms. Lastly, a big amount of data in industry require effective and high-speed processing.\\
\indent Attempting to address the challenges, traditional time series prediction methods that will be mentioned in Section~\ref{previous work} are useful but not accurate enough. Besides, efficiency should also be taken into consideration. Thereby, our work aims to predict production or flow more accurately, from the aspect of industrial time series data with those aforementioned attributes. Our main contribution consist in improving the traditional LSTM algorithm for higher accuracy mainly in two aspects. First, we firstly consider the periodicity of industrial data, so that the data can fit the model better and thus promoting the accuracy. Second, we consider high dependencies involving multiple variables by transforming data. We evaluate the ameliorated algorithm by analyzing theoretically and do comparison experiment as well.
\paragraph*{Outline}
The remainder of this article is organized as follows. Section~\ref{previous work} presents related work. Section~\ref{problem definition} gives an account of how the problem is defined. Section~\ref{LSTM based on multivariate tuning} describes the modules that constitute the proposed algorithm. Section~\ref{algo descrpt & analysis} overview the algorithm and analyze it with respect to time and spatial complexity. Then, our results are evaluated with experiments in Section~\ref{experiment eval}. Finally, Section~\ref{conclusion} draws the conclusions.

\section{Related Work}\label{previous work} 
\paragraph{Time Series Prediction} The time series prediction\cite{martinetz1993neural, brockwell2002introduction} analyzes and uses the features of data from specific past time periods to predict the characteristics of future data. The construction of the time series models is closely related to the order that the happening time of events follows, and it is more complicated than that of normal regression predictions. Time series problems can be partitioned into several types according to the following factors: whether the time series is repetitive, whether the change factors are independent, whether the dimensionality reduction or the dimensionality ascending is used, whether there are complex historical dependencies, and so on.

\indent Since 1991, the backpropagation neural network\cite{weigend1991back, wong1991time} has already been applied to time series prediction models. With the development in this field, the improved neural network algorithms have been more widely used to forecast time series. The most common tool of all these models is the recurrent neural network (RNN)\cite{cai2007time}, which with memories, can preserve the results of previous calculations. The hidden layer in the RNN considers both the current input and the results of the last hidden layer, unlike traditional neural networks in which there is no dependency between the calculation results. However, in order to achieve long-term memory, the RNN model requires a significant amount of model training time. Thus, the long short-term memory network (LSTM) is proposed to shorten the training time\cite{ma2015long}. Besides, various time series prediction algorithms based on support vector machines\cite{sapankevych2009time} have also been put forward to deal with different time series prediction problems. In industry, time series predictions are also increasingly used to predict problems, such as renewable energy prediction\cite{doucoure2016time}, financial time series prediction\cite{tenti1996forecasting}, and clinical medical data prediction\cite{che2018recurrent}, and have achieved outstanding results.\\
\indent Among all these algorithms for predicting time series, the random forest, the backpropagation neural network, the convolutional neural network (CNN), and the long short-term memory (LSTM) are common, which thus will be compared with the algorithm proposed later in the paper.

\paragraph{Long Short-term Memory} RNN is a general term for time recurrent neural networks and structural recurrent neural networks including LSTM. In this paper, what we focus on is the former one. Unlike common feedforward neural networks that accept specific structural inputs, RNN passes the state of the past to its own network iteratively, so the input of the RNN can be a time series structure.\\
\indent The RNN has several steps. First, denote the input as $x= (x_t, \cdots , x_2, x_1)$ and the hidden state as $h_t = f(x_t, h_{t-1})$. The hidden state updates with time. Then, after inputting the new data, the hidden state $h_{t-1}$ is transformed to $h_t$ with respect to $x_t$. In every update, the weights of the previous series are weakened, so that the time correlation is presented. Note that the hidden state $f(x, h)$ is a nonlinear function.\\
\indent The LSTM can learn and memorize long-term information, with its structure resembling that of the RNN. Yet the main difference lies in that the RNN has a single neural network layer, while the LSTM has 4 in every unit. Besides, the LSTM interacts in a special way, as is shown in Figure 1.\\
\begin{figure}[ht]
	\centering
	\includegraphics[width=6.2cm]{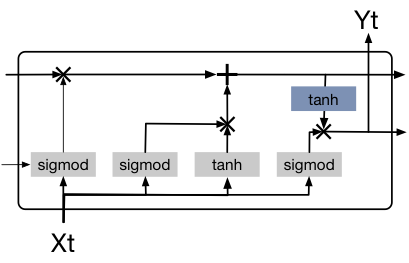}
	\caption{LSTM Neuron}
	\label{fig0}
\end{figure}
\indent The core idea of the LSTM revolves around the state of the unit, which is designed with a structure called a gate to modify the state by adding or removing certain information. Besides, the gate can filter the input and contains layers such as sigmoid pointwise. Sigmoid controls the amount of the input data allowed to pass, whose value is between 0 and 1. Obviously, 0 blocks all information, while 1 allows any amount to pass. The LSTM uses a three-door structure in each unit to maintain the state of the unit.\\
\indent The LSTM first determines what information will be excluded from the state of the unit. This operation is implemented by a sigmoid layer called a forget gate. The forget gate read the last hidden state $h_{t-1}t$ and the current input $x_t$, and output a value $f_t$ between 0 and 1. Then, it determines what information is reserved in the current unit. This operation includes two parts. The first part uses the sigmoid layer to decide the updating target and then set the current data $x_t$ as the input. The second part uses the tanh layer to construct the vector $\boldsymbol{C_t'}$. $x_t$ and $\boldsymbol{C_t'}$ will play a role in the subsequent unit status updates.\\
\indent The $\boldsymbol{C_{t-1}}$ updated in the forget gate will take part in the following calculation to attain the current state $\boldsymbol{C_t}$:
\begin{equation}
\boldsymbol{C_t} = \boldsymbol{C_{t-1}}\times f_t + i_t \times \boldsymbol{C_t'}
\end{equation}
\indent Lastly, the LSTM determines the output for the next moment. The $\boldsymbol{C_t}$ calculated by the previous step will have two output directions, one of which is directly connected to the unit at the next moment as its input data, and the other requires a filtering process. The process first pass the current data $x_t$ and the hidden state from last moment $h_{t-1}$ into the sigmoid layer as the input, and we have the output $vp$. Then, the result of the tanh layer after passing the unit state $\boldsymbol{C_t}$ is multiplied with $vp$, and we get the current hidden state. Finally, the product is passed in the unit of the next moment.\\

\section{Problem Definition}\label{problem definition}

\indent In this section, we define our problem as to find a model with the relevant parameters of the device as input. For a certain input together with predictive values, there exists a function that can map the input to the predictive values in the model.
\indent In practice, the input, production or flow, is industrial time series data $\boldsymbol{X} = [\boldsymbol{x}_1 \  \boldsymbol{x}_2 \  \ldots \  \boldsymbol{x}_m] \in \boldsymbol{R}^{d\times m}$, and the output is the flow predictive model $M$. To formalize this problem, given the time series data acquired from an industrial device $\boldsymbol{X} = [\boldsymbol{x}_1 \  \boldsymbol{x}_2 \  \ldots \  \boldsymbol{x}_m] \in \boldsymbol{R}^{d\times m}$, the true values $\boldsymbol{Y} = [y_1 \  y_2 \  \ldots \  y_n]^T$, for any column vector of which the correlations between data $t-1, t, t+1$ are unknown, and $f(\cdot)$ is a function in $M$ that calculates the predicted value. Thus, the problem is to find:
\begin{equation}
\mathop{\arg\min}_{M}  RSME(M(\boldsymbol{X}, f(\cdot)), \boldsymbol{Y})
\end{equation}

\indent Our model is chosen according to the characters of time series data due to the data type of matrix $X$. As the prediction target is serial data with complex mechanisms and nonlinear dependency relationship, rendering the model more sensitive to the prediction accuracy, a well-predicting neural network algorithm is needed. For example, the backpropagation and the long short-term memory (LSTM) are both available in this situation.\\
\indent In this paper, LSTM, a special recurrent neural network (RNN), is adopted, considering the fact that industrial data have a problem of long-term dependencies and LSTM is capable of solving it. Although its training processing is relatively slow and training dimensions high, the merits of LSTM that it is powerful in sequence modeling, able to store previous information and to fit nonlinearities, make LSTM suitable for predicting time series data.\\
\indent However, concerning industrial data, using LSTM directly is difficult to acquire the step size in each iteration according to simple inferences, because industrial data have imperceptible periodicity. Besides, the real value of production or flow not only fluctuates with time, but it has high correlations with varying parameters. Therefore, according to such data traits, we propose an LSTM algorithm based on multi-variable tuning.
\section{LSTM Based on Multivariate Tuning}\label{LSTM based on multivariate tuning}
\subsection{Overview}
\indent The LSTM algorithm based on multivariate tuning has three modules, including a data conversion module, an LSTM modeling module, and a tuning module, as is shown in Figure 2. The data conversion module changes time series data into supervised learning sequences and finds the variable sets with which the predictive value $\boldsymbol{Y}$  has the highest regression coefficient. The LSTM modeling module forms an LSTM network, connecting multiple LSTM perceptrons. The tuning module adjusts parameters according to root-mean-square deviations (RMSE) in each iteration and then pass the altered data to the data conversion model for training again so that after continuous iterations, the approximate optimal solution can be found.
\begin{figure}[ht]
	\centering
	\includegraphics[width=8cm]{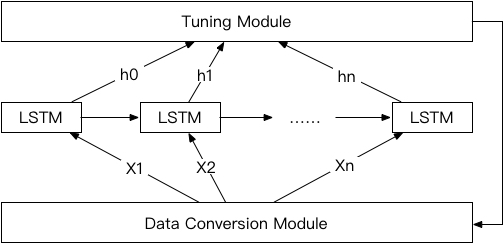}
	\caption{LSTM based on multivariate tuning}
	\label{fig1}
\end{figure}
\subsection{Data Conversion Module}
\indent The data conversion module, aiming at converting time series data to supervised learning sequences for later use, and it solves the multivariate problem with high dimensions by utilizing periodicity of the data and transforming rows of data to one row. This module includes two operations, data preprocessing and data converting.\\
\indent The first data preprocessing operation reduces the dimensionality of the input data to approach its goal, with three steps. First, according to the input variable $\boldsymbol{X} = [\boldsymbol{x}_1 \  \boldsymbol{x}_2 \  \ldots \  \boldsymbol{x}_m] \in \boldsymbol{R}^{d\times m}$, using the method of random decision forests, the correlation coefficients $C = [c_1 \  c_2 \  \ldots c_m]$ can be calculated. Second, the correlation coefficients are ranked in order $c_1\ge c_2\ge c_3\ge \ldots \ge c_m$. Third, the result of this step is variables $\boldsymbol{X'} = [\boldsymbol{x'}_1 \  \boldsymbol{x'}_2 \  \ldots \  \boldsymbol{x'}_m] \in \boldsymbol{R}^{d\times m}$, whose sum of correlation coefficients are greater than 95\%, namely $\sum_i c_i \ge 95\%$. In this way, based on random forest analysis of the data, the weight of each variable is estimated, and the data's dimensionality is reduced.  \\
\indent The second data converting operation is aimed at merging several rows to one, changing time series data into supervised learning sequences according the periodicity of the data. Thus, the problem defined in this operation is, given a positive integer $n$ and $\boldsymbol{X'} = [\boldsymbol{x'}_1 \  \boldsymbol{x'}_2 \  \ldots \  \boldsymbol{x'}_m] \in \boldsymbol{R}^{d\times m}$, to select all data from the $(i-n)$th, $(i-n+1)$th, \ldots, $(i-1)$th row where $i > n$, select the first $m-1$ data in the $i$th row as independent variables $\textit{X}$ and the $m$th datum as the dependent variable $\textit{Y}$, namely the predictive value.\\
\indent Algorithm 1 outlines the process to transform time series data. In the first step, weights or importance are attained using random forest trees (lines1 to 3). Subsequently, the data with the sum of weights greater than 95\% are selected (lines 4 to 12). Finally, $n$ rows of data are transformed into one row (lines 13 to 17). For example, when $n = 3$, every three rows of data are merged into a new row, the last value of which is the predictive value $\boldsymbol{Y}$, as presented in Figure 3.
\begin{algorithm}
	\caption{data\_transform($data_D, n$) in Data Conversion Module}
	\label{alg1}
	\hspace*{0.02in} {\bf Input:} the input dataset $data_D$, a positive integer $n$\\
	\hspace*{0.02in} {\bf Output:} the output dataset $data_T$
	\begin{algorithmic}[1]
		\STATE $forest \leftarrow$ randomforestregressor($data_D$)
		\STATE $importances \leftarrow$ importance($forest$)
		\STATE sort($importances$)
		\STATE $threshold \leftarrow 0$
		\STATE $colSet \leftarrow$ NULL
		\FOR{each $p \in importances$}
		\STATE $colSet \leftarrow colSet.add(p.name)$
		\STATE $threshold \leftarrow threshold +p.importance$
		\IF{$threshold > 0.95$}
		\STATE break
		\ENDIF
		\ENDFOR
		\STATE $data_D \leftarrow data_D[colSet]$
		\STATE $row\_size \leftarrow$ shape($data_D$)
		\STATE $new\_row\_siez \leftarrow$ int($row\_size/n$)
		\STATE $data_T \leftarrow data_D.ix[0 : new\_row\_size*n, :]$
		\STATE reshape($data_T$)
		\RETURN $data_T$
	\end{algorithmic}
\end{algorithm}
\begin{figure}[ht]
	\centering
	\includegraphics[width=8cm]{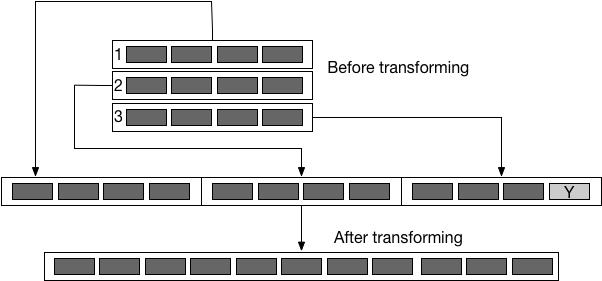}
	\caption{Data Transforming}
	\label{fig2}
\end{figure}

\subsection{LSTM Modeling Module}
\indent The LSTM modeling module connects multiple LSTM perceptrons to form an LSTM network, aiming to preserve the impact of early emerging information and connect it to the present task. LSTM perceptrons control output states, using forget gates and input gates, where forget gates determine how many output states from previous units are retained, while output gates determine how many current states are retained. Thereby, LSTM perceptrons can achieve its goal.\\
\indent A single-layer LSTM network is implemented in this module to reduce the modeling time and guarantee the predictive effect, given the timeliness of industrial time series data and the fact that the model needs multiple iterations to find the optimal parameters. The model adopts the traditional 3-layer LSTM network structure. The first layer is the input layer, in which the number of neurons is equal to the data dimension of each input. The second layer is the hidden layer, whose number of neurons is determined by experimental results. The third layer is the output layer. Because the problem solved is a single-valued prediction problem, the output layer is a single neuron.\\
\indent For example, when the input vector has 3 variables, LSTM has 4 nuerons, as depicted in Figure 4.
\begin{figure}[ht]
	\centering
	\includegraphics[width=8cm]{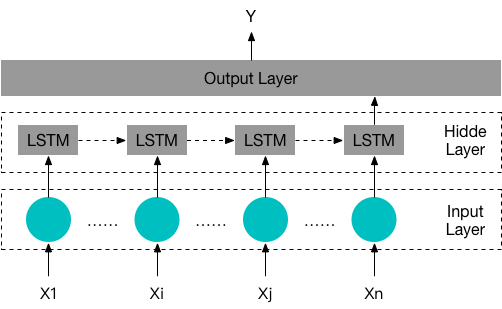}
	\caption{Single-Layer LSTM Network Structure}
	\label{fig3}
\end{figure}

\subsection{Tuning Module}
\indent The tuning module is an essential part, where time periodicity is taken into consideration, even though this trait in industry is difficult to observe. By analyzing the cycles, the accuracy can be greatly promoted. This module is composed of two parts, periodic analysis and iterative tuning.\\
\indent The first periodic analysis part calculates the periodic law of the predictive value $\boldsymbol{Y}$. First, in order to increase the accuracy of calculating the period value, the predicted values should be normalized, that is, the data should be converted into the $(0, 1)$ interval. Because the definition of periodicity indicates that the data change following a cyclical law, the features of data fluctuation can be utilized to record the point where each value changes from negative to positive and vice versa. In other words, it is the position of the data passing through the zero point in the time series data that is recorded. Then, the differences of the recorded positions are set as the representative period value, and the first 5 periods with the smallest period value are taken as the number of iterations.\\
\indent The algorithm in this part is implemented leveraging the minimum heap. Algorithm 2 outlines the process more specifically. In the first step, the data is regularized (line 1), and then in the second step, the period counter $count$ and the heap are initialized (line 2 to 3). Each iterative operation of the third step (line 4 to 18) increases $count$ by $1$ if the positive or negative of the value $y$ does not change, or stores the counter $count$ in the heap if it changes. Note that the size of the minimum heap is set to 5.\\
\begin{algorithm}
	\caption{cycle($data_D.y$) in Tuning Module}
	\label{alg2}
	\hspace*{0.02in} {\bf Input:} the predictive column $data_D.y$ of the dataset $data_D$\\
	\hspace*{0.02in} {\bf Output:} the period value $stepSet$
	\begin{algorithmic}[1]
		\STATE $ySet \leftarrow$ regular($data_D.y$)
		\STATE $count \leftarrow 0$
		\STATE $stepSet \leftarrow$ heap()
		\FOR{each $y \in ySet$}
		\IF{$y > 0$}
		\STATE $count \leftarrow count + 1$
		\ELSE
		\IF {$count = 0$}
		\STATE continue
		\ELSE
		\STATE $stepSet \leftarrow stepSet$.push($count$)
		\STATE $count \leftarrow 0$
		\IF{size($stepSet$) $>5$}
		\STATE $stepSet \leftarrow stepSet$.pop()
		\ENDIF
		\ENDIF
		\ENDIF
		\ENDFOR
		\RETURN $stepSet$
	\end{algorithmic}
\end{algorithm}
\indent The second iterative tuning part is based on the training to obtain the training model $M$ and the root mean square error $RMSE$ to determine whether to continue the iteration until the minimum $RMSE$ is found by changing the size of $n$ in data\_transform($data_D,n$). Given the uncertainty of the optimal solution of $n$ ($n<\frac{m}{2}$), and the fact that $n$ and RSME do not satisfy the linear relationship, the periodic value sequence obtained by the periodic analysis part is used as the input of $n$ for iterative calculation. After the iterations, the minimal $RSME$ calculated is the approximate optimal solution of the algorithm. It can be inferred that there is no overfitting from the fact that the result in the algorithm is the calculation result of the fitting model of the first 5 periods. Although we obtained the approximate optimal solution instead of the global solution, the risk of overfitting is therefore reduced to some extent.

\section{Algorithm Description and Analysis}\label{algo descrpt & analysis}
\subsection{Description}
\indent After sorting the above modules, the calculation flow of the algorithm is obtained, the LSTM based on multivariate tuning is represented in Algorithm 3 as a whole. First, we need to process the data to obtain the periodic value of the predictive value $y$. Second, the data is transformed according to the periodic value. Third, the LSTM algorithm is then used to model. This third process is iterated until the first 5 minimal periods are trained.\\
\begin{algorithm}
	\caption{optimizedLSTM($data_D$)}
	\label{alg3}
	\hspace*{0.02in} {\bf Input:} the dataset $data_D$\\
	\hspace*{0.02in} {\bf Output:} the optimal model $M_f$
	\begin{algorithmic}[1]
		\STATE $stepSet[]$ $\leftarrow$ cycle($data_D.y$)
		\STATE $count \leftarrow 0$
		\STATE $data_T \leftarrow$ data\_transform($data_D.stepSet[count]$)
		\STATE $m \leftarrow \infty$
		\WHILE{$count < 5$}
		\STATE $M \leftarrow$ LSTM($data_T$)
		\IF{$m > M.RSME$}
		\STATE $m \leftarrow M.RSME$
		\STATE $M_f \leftarrow M$
		\ENDIF
		\STATE $count \leftarrow count + 1$
		\ENDWHILE
		\RETURN $M_f$
	\end{algorithmic}
\end{algorithm}
\indent First, the cycle sequence is acquired (line 1). Second, the iteration parameter is initialized (line 2). Then, the data conversion module transforms the data (line 3). Last, after iterations of training with the revised LSTM model, the final optimized model $M_f$ is obtained. For better understanding, the algorithm flow chart is shown in Figure 5.
\begin{figure}[ht]
	\centering
	\includegraphics[width=8cm]{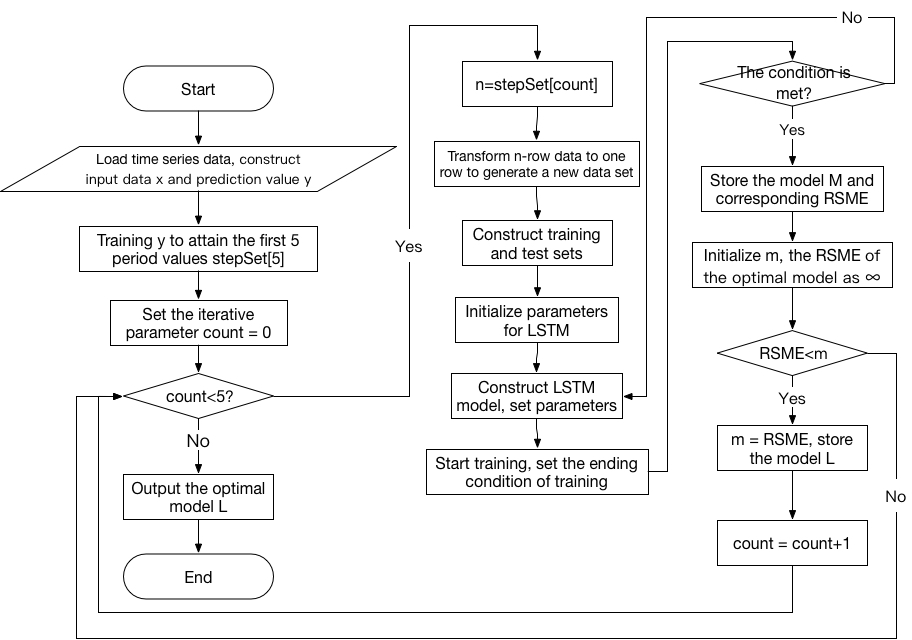}
	\caption{LSTM Based on Multivariate TuningFlow Chart}
	\label{fig4}
\end{figure}
\subsection{Analysis}
The analysis toward the LSTM based on multivariate tuning lies in time and spatial complexity in the training process. It shows that out work is superior to some other algorithms.
\paragraph{Time Complexity}
The LSTM based on multivariate tuning leverages its periodicity to iterate. To start with, let us assume that the size of the training data is $N$. Because the operation of obtaining the periodic value only goes through one iteration, its time complexity is $O(N)$. Let us assume that there are $L$ iterations in the iterative tuning part, each of which generates an LSTM network. For each LSTM network, let us assume it has $C$ iterations, a sequence in each iteration has a length of $S$, and the size of the input data is $W$. First, the process of calculating hidden states of the input data and forward propagation has a time complexity of $S\cdot W$. Then, the adjustment of coefficients according to the Adam algorithm\cite{kingma2014adam} has a time complexity of $S\cdot 1$. Thus, the time complexity of going through an $S$-long sequence is $S\cdot (W+1)$. Lastly, because there are $C$ iterations within which there are $\lceil \frac{N}{S} \rceil$ $S$-long sequences, the time complexity of training an LSTM network is $O(S\cdot (W+1)\cdot \lceil \frac{N}{S} \rceil \cdot C) = O(C\cdot N\cdot W)$. In conclusion, for the overall training with $L$ iterations, the total time complexity is $O(N)+L\cdot O(C\cdot N\cdot W) = O(N+L\cdot C\cdot N\cdot W)$.
\paragraph{Spatial Complexity}
The size of the data involved is $N$. In the training, the input data has $I$ dimensions; the hidden layer has $H$ nodes, and the output data is $O$-dimensional. Thereby, the cost of the transformation from the input layer to the hidden layer is $H\cdot I$; the spatial cost of the states changing in the hidden layer is $H^2$; the cost of the transformation from the hidden layer to the output layer is $O\cdot H$. Besides, the parameters in the hidden layer costs $H\cdot 1$, and the parameters in the output layer costs $O\cdot 1$, which is used to store the parameter of the optimal model. Let us assume that a sequence in each iteration has a length of $S$,  and thus the forward propagation of an $S$-long sequence when going through the input layer, hidden layer and the output layer costs $S\cdot I\cdot 1$, $S\cdot H\cdot 1$ and $S\cdot O\cdot 1$ respectively. Moreover, the backpropagation, using the Adam algorithm to adjust coefficience, the spatial costs are respectively, $H\cdot I$, $H^2$, $O\cdot H$, $H\cdot 1$, and $O\cdot 1$. Therefore, the overall spatial cost of a single LSTM modeling is:
\begin{equation}
2HI + 2H^2 + 2OH + 2H + 2O + SI + SH + SO
\end{equation}
\indent Considering there are $L$ iterations, in each of which an LSTM network is generated  and one LSTM network has $\lceil \frac{N}{S} \rceil$ sequences with a length of $S$ to train, the total spatial cost is: $LN + L(2\lceil \frac{N}{S} \rceil + 1)(HI + H^2 + OJ + H + O) + \\L\lceil \frac{N}{S} \rceil(SI + SH + 2SO)$\\
\indent In conclusion, the spatial complexity is:
\begin{equation}
O(LN + L\lceil \frac{N}{S} \rceil H^2)
\end{equation}

\section{Experimental Evaluation}\label{experiment eval}
\subsection{Evaluation Method for Prediction Problem}
\indent We select the root-mean-square error (RSME) to appraise effects of the algorithms. First of all, two methods have to be explicitly pointed out. Both RSME and R-squared (R$^2$) are effective to measure the prediction result. Differnt from RSME, a quantitative evaluation, R$^2$ compares the prediction value with the situation where only the mean is used, and then identify how much better is the prediction. Besides, R$^2$ generally lies in the $(0, 1)$ interval. However, RSME is more sensitive to outliers. When some prediction value and the true value differ greatly, the RSME will increase. Therefore, RSME is more suitable for a circumstance demanding high precision measurements where industrial big data are included, and RSME is chosen to evaluate the algorithm:
\begin{equation}
RSME (X, f(\cdot)) = \sqrt{\frac{1}{m}\sum_{i=1}^{m}(f(x_i)-y_i)^2}
\end{equation}
\subsection{Experimental Condition}
\indent Table 1 shows the experimental condition. The data used in the experiments is from the industrial boiler dataset of the supervisor information system(SIS) in a Harbin electric power plant. The data include time, flow, pressure, and temperature, adding up to 70 dimensions and more than 400,000 pieces. The experimental condition is shown in the table.\\
\linespread{1.5}
\begin{table}[ht]
	\caption{Environmental Condition}
	\scriptsize
	\label{tbl1}
	\centering
	\begin{tabular}{c|c}
		\hline
		\multirow{2}{3.5cm}{Machine Configuratio} & \multirow{2}{4cm}{2.7GHz Intel Core i5,\\8GB 1867 MHz DDR3}\\
		\\
		\hline
		\multirow{1}{3.5cm}{Environment} & \multirow{1}{4cm}{Python 3.6.0, Tensorflow}\\
		\hline
		\multirow{1}{3.5cm}{Dataset} & \multirow{1}{4cm}{Industrial Boiler Dataset}\\
		\hline
	\end{tabular}
\end{table}
\indent The experiment reduces the experimental error by multiple measurements and averaging. Here, the holdout method is used to verify the algorithm, that is, the data set is randomly divided into a training set and a test set. In this experiment, the training set accounts for 2/3 of the data set, and the test set accounts for 1/3 of the data set for verification.\\
\indent The experiment optimizes the parameters of the LSTM based on multivariate tuning and then compares the result of the optimized LSTM with other algorithms shown in Table 2.\\
\begin{table}[ht]
	\caption{Comparison Experiment Design}
	\label{tbl2}
	\scriptsize
	\centering
	\begin{tabular}{c|c}
		\hline
		Experiment Label & Adopted Algorithm\\
		\hline
		Comparative Experiment \uppercase\expandafter{\romannumeral1} & Random Forest\\
		Comparative Experiment \uppercase\expandafter{\romannumeral2} & Backropagation\\
		Comparative Experiment \uppercase\expandafter{\romannumeral3} & CNN\\
		Comparative Experiment \uppercase\expandafter{\romannumeral4} & LSTM\\
		Experiment & Optimized LSTM\\
		\hline
	\end{tabular}
\end{table}
\subsection{Tuning Process}
\indent The experiment was conducted using the TensorFlow, a comprehensive and extensible framework. Through neural network theory analysis, it can be found that the parameters affecting the performance of the prediction method mainly include the number of iterations, the learning rate, the length of the iterative sequence, and the number of nodes in the hidden layer. Since the experiment uses the Adam algorithm to adaptively adjust the learning rate, the algorithm is optimized from the number of iterations, the length of the iterative sequence, and the number of nodes in the hidden layer.\\
\indent To optimize the length of the iterative sequence, the performances of the prediction models are recorded at sequence lengths of 10, 50, 100, 250, 500, and 1000. The experiments shown in Table 3 were designed. In the experiments, the numbers of hidden layer nodes are all  50, and the preset numbers of iterations are all 50. It can be seen by comparison that the length of the sequence has some but small influence on the experimental results. However, an iterative sequence length of 500 is regarded as performing well.\\
\begin{table}[ht]
	\caption{Parameters for Optimal Iterative Length }
	\centering
	\scriptsize
	\label{tbl3}
	\begin{tabular}{c c c c c}
		\hline
		\multirow{2}{1cm}{Exprmnt\\ Label}&\multirow{2}{1.4cm}{Sequence Length} &\multirow{2}{1.6cm}{\# of Hdn Lyr Nodes}& \multirow{2}{1.4cm}{\# of Iterations} & \multirow{2}{1cm}{RSME}\\
		\\
		\hline
		1 & 10 & 50 & 50 & 22.83\\
		2 & 50 & 50 & 50 & 17.95\\
		3 & 100 & 50 & 50 & 19.20\\
		4 & 250 & 50 & 50 & 14.64\\
		5 & 500 & 50 & 50 & 11.49\\
		6 & 1000 & 50 & 50 & 13.85\\
		\hline
	\end{tabular}
\end{table}
\indent To optimize the number of hidden layer nodes, the performances of the prediction models with the number of hidden layer nodes set to 10, 20, 50, and 100 are recorded. The experiments are designed based on the result of the optimal sequence experiment, and thus in these experiments, the lengths of the iterative sequence are all set to 500, and the preset number of iterations is 50. It can be seen from Table 4 and Figure 6 that the RMSE is the smallest when the number of the hidden layer nodes is 200, which means that for a single-layer LSTM network, the more the number of hidden layer nodes is, the better the experimental result is. However, when the number of hidden layer nodes reaches 200, the optimization rate of the experimental results is slowed down. Therefore, when the number of hidden layer nodes is 100, it can achieve a better experimental result, so the number of hidden layer nodes is tentatively set to 100.\\
\begin{table}[ht]
	\caption{Parameters for Number of Hidden Layer Nodes}
	\label{tbl4}
	\centering
	\scriptsize
	\begin{tabular}{c c c c c}
		\hline
		\multirow{2}{1cm}{Exprmnt\\ Label}&\multirow{2}{1.4cm}{Sequence Length} &\multirow{2}{1.6cm}{\# of Hdn Lyr Nodes}& \multirow{2}{1.4cm}{\# of Iterations} & \multirow{2}{1cm}{RSME}\\
		\\
		\hline
		7 & 500 & 10 & 50 & 79.85\\
		8 & 500 & 20 & 50 & 18.98\\
		5 & 500 & 50 & 50 & 11.49\\
		9 & 500 & 100 & 50 & 9.13\\
		10 & 500 & 200 & 50 & 8.69\\
		\hline
	\end{tabular}
\end{table}
\indent To optimize the number of iterations, the experimental results of different iterations of 10, 50, 100, and 500 are recorded. The following experimental results are designed based on the above experimental results. In this set of experiments, the lengths of each iteration sequence are 500, and the numbers of hidden layers are 100. It can be seen from Table 5 and Figure 7 that when the number of iterations is 50, 100 and 500 respectively, the final experimental results are equivalent, but the smaller the number of iterations is, the less time it takes, so the number of iterations will be It is tentatively set to 50 times.\\
\begin{table}[ht]
	\caption{Parameters for Number of Iterations}
	\label{tbl5}
	\centering
	\scriptsize
	\begin{tabular}{c c c c c}
		\hline
		\multirow{2}{1cm}{Exprmnt\\ Label}&\multirow{2}{1.4cm}{Sequence Length} &\multirow{2}{1.6cm}{\# of Hdn Lyr Nodes}& \multirow{2}{1.4cm}{\# of Iterations} & \multirow{2}{1cm}{RSME}\\
		\\
		\hline
		11 & 500 & 100 & 10 & 31.93\\
		9 & 500 & 100 & 50 & 9.13\\
		12 & 500 & 100 & 100 & 9.47\\
		13 & 500 & 100 & 500 & 9.96\\
		\hline
	\end{tabular}
\end{table}

\begin{figure*}[ht]
	\centering
	\subfigure[\#Hidenodes VS. RSME]{
		\includegraphics[width=0.3\linewidth]{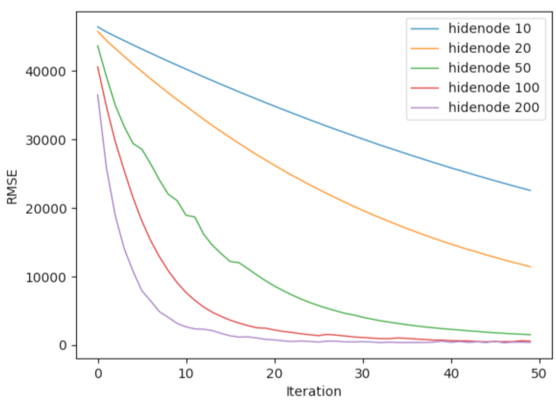}
		\label{fig5}
	}
	\subfigure[\#Iterations VS. RSME]{
		\includegraphics[width=0.3\linewidth]{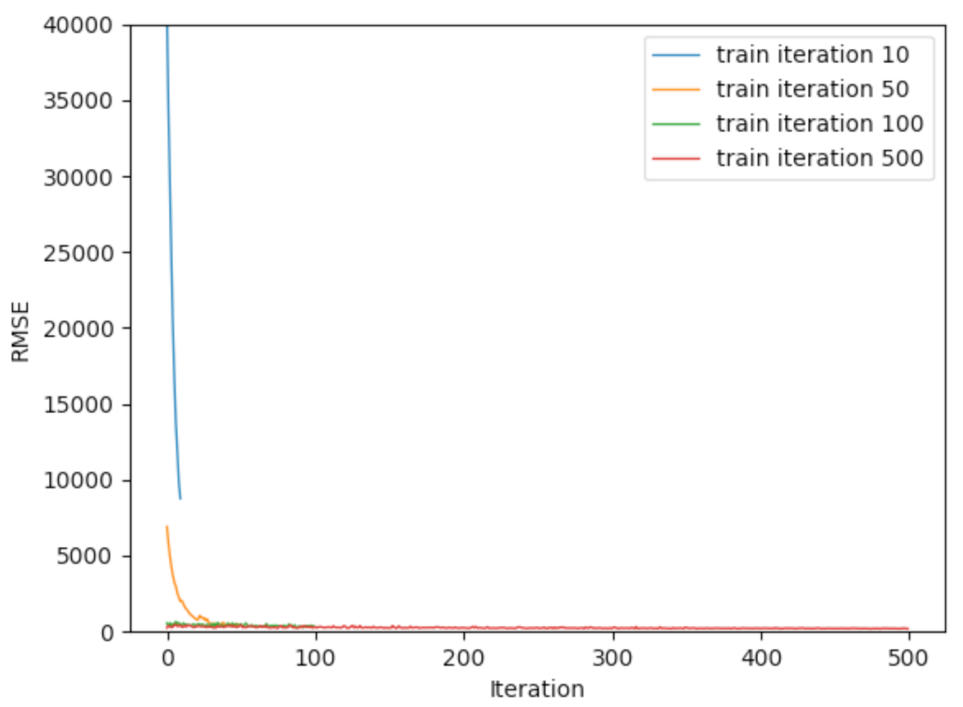}
		\label{fig6}
	}
	\subfigure[Prediction Values VS. Actual Values]{
		\includegraphics[width=0.3\linewidth]{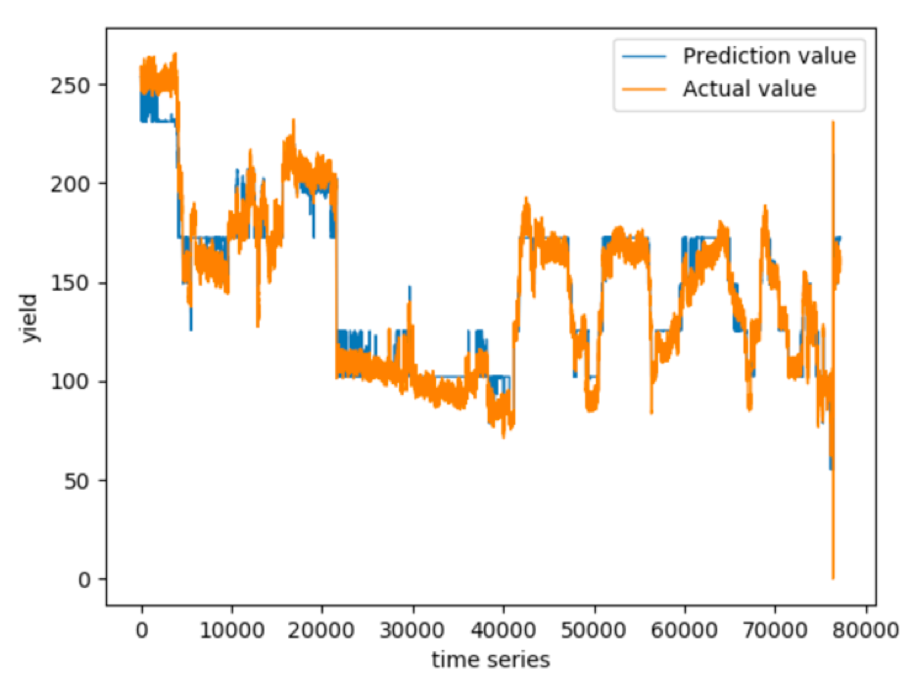}
		\label{fig7}
	}
	\subfigure[LSTM RSME VS. Iterations]{
		\includegraphics[width=0.3\linewidth]{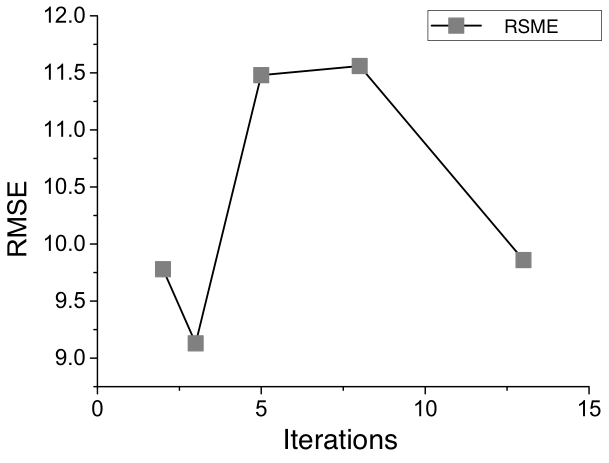}
		\label{fig8}
	}
	\subfigure[Loss Rate]{
		\includegraphics[width=0.3\linewidth]{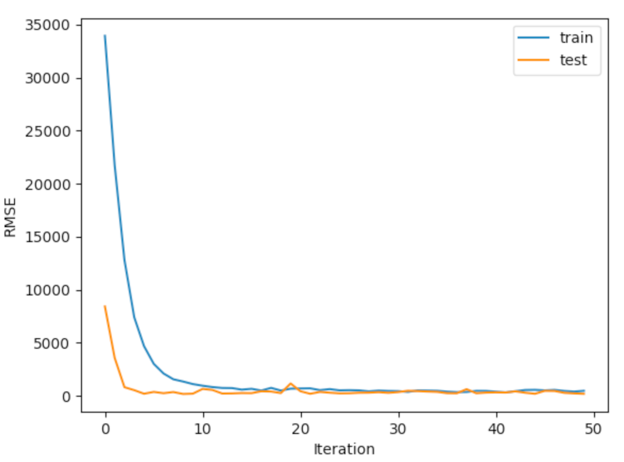}
		\label{fig9}
	}
	\subfigure[Comparisons]{
		\includegraphics[width=0.3\linewidth]{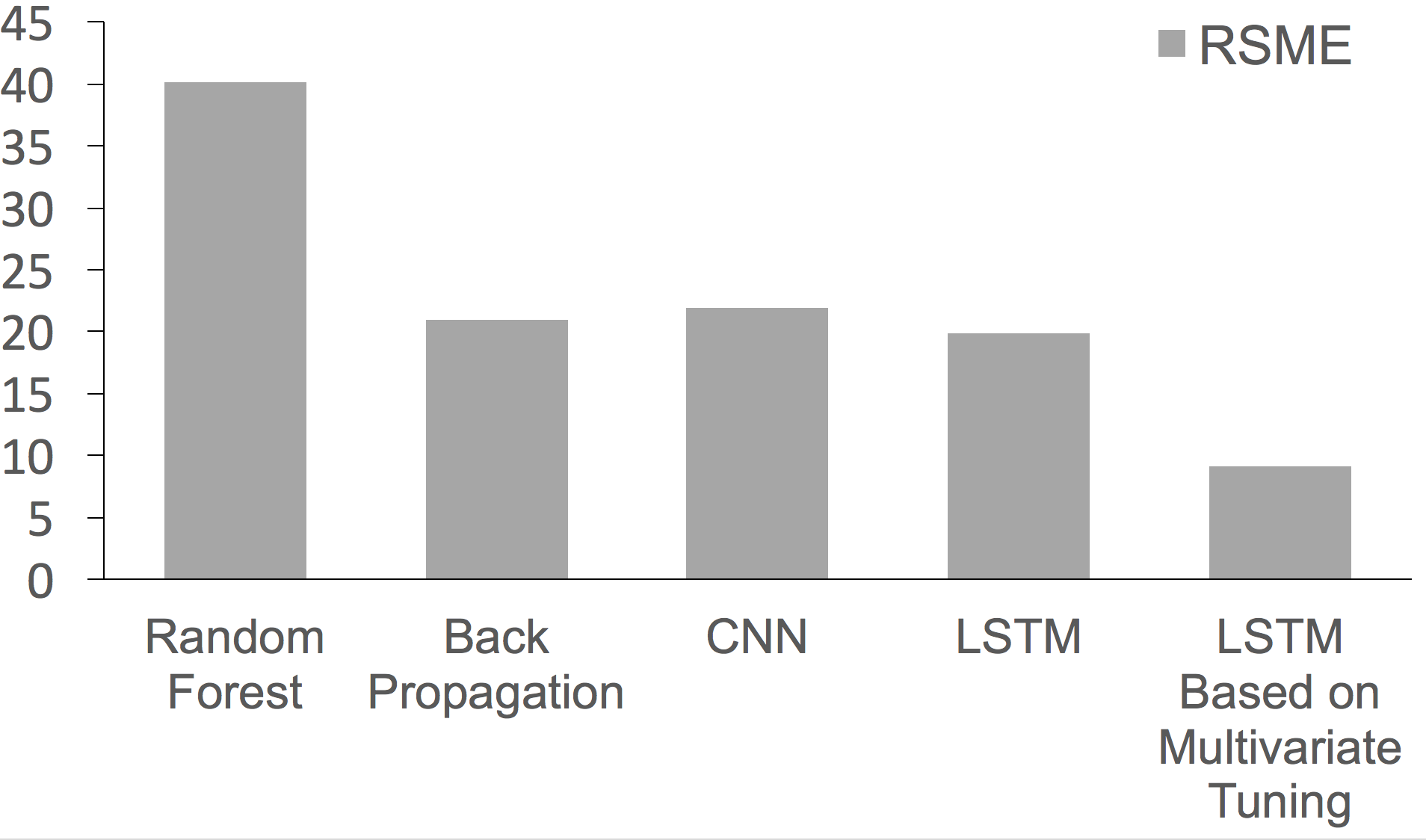}
		\label{fig10}
	}
	\caption{Experimental Results}
\end{figure*}

\indent Based on the above experiments, the experimental parameters of the optimal results are determined. From the experimental results, when the length of each iteration is 500, the number of hidden layers is 100, and the number of iterations is 50, the experimental results are optimal, as shown in Table 6. Due to the limitations of parameter tuning, it does not rule out the emergence of more optimized experimental results. The prediction results under optimal parameters are shown in Figure 8.\\
\begin{table}[ht]
	\caption{Parameters for Optimal results}
	\label{tbl6}
	\centering
	\scriptsize
	\begin{tabular}{c c c c c}
		\hline
		\multirow{2}{1cm}{Exprmnt\\ Label}&\multirow{2}{1.4cm}{Sequence Length} &\multirow{2}{1.6cm}{\# of Hdn Lyr Nodes}& \multirow{2}{1.4cm}{\# of Iterations} & \multirow{2}{1cm}{RSME}\\
		\\
		\hline
		9 & 500 & 100 & 50 & 9.13\\
		\hline
	\end{tabular}
\end{table}

\indent The algorithm is analyzed when the parameters are optimal. The iterative process of the LSTM based on multivariate tuning is shown in Figure 9. The first five values with the smallest period value are selected for iteration. It can be found that the RMSE does not increase with the increase of $n$, and it exhibits a nonlinear trend. When $n$ is 3, the local minimum value of 9.13 is obtained which currently the best training results.\\

\indent It can be proved that the model is not overfitting. When $n$ is 3, the loss rate curve of the LSTM based on multivariate tuning is shown in Figure 10. Comparing the loss rate curve after training the training set and test set,  we can find that the loss rate of the two tends to be consistent after a period of time, and the test set loss rate is not higher than the training set loss rate, which proves that the model is not overfitting.\\
\subsection{Comparison Experiment Result}
\indent According to the comparison of the control experiments in Section 5.2, the experimental results of the random forest, the backpropagation neural network, the convolutional neural network (CNN), the long short-term memory (LSTM), and the LSTM based on multivariate tuning are compared. The experimental results are shown in Table 7.  From the values of RMSE, the results of the random forest experiment are larger than those of backpropagation neural network. The experimental results of the backpropagation neural network and convolutional neural network are nearly the same. The original LSTM algorithm is better than the experimental results of the above three algorithms. In Experiment \uppercase\expandafter{\romannumeral1} to \uppercase\expandafter{\romannumeral4}, the results of the LSTM algorithm are the best. Thus, it is valuable to choose the LSTM algorithm for optimization. The experimental results prove that the LSTM algorithm is more suitable for time series prediction. Compared with the random forest with an RMSE of 40.2, the original LSTM algorithm has an RMSE of 19.87, which is 50\% less.\\
\indent Compared with the LSTM algorithm before and after optimization, the LSTM algorithm with multivariate tuning is the best prediction algorithm whose RMSE is 9.13. However, that of the original LSTM is 19.87, and thereby the LSTM based on multivariate tuning is optimized by 54.05\%. Among them, the LSTM algorithm can effectively target time series data mainly because of its long-term and short-term memory, and the improved LSTM algorithm can make the model more optimized, which is the best experimental algorithm. The comparison results are shown in Figure 11.\\
\begin{table}[ht]
	\caption{Comparison Experiment Design}
	\label{tbl7}
	\centering
	\scriptsize
	\begin{tabular}{c|c|c}
		\hline
		Experiment Label & Adopted Algorithm & RSME\\
		\hline
		Comparison Exp. \uppercase\expandafter{\romannumeral1} & Random Forest & 40.21\\
		Comparison Exp. \uppercase\expandafter{\romannumeral2} & Backropagation & 20.90\\
		Comparison Exp. \uppercase\expandafter{\romannumeral3} & CNN & 21.97\\
		Comparison Exp. \uppercase\expandafter{\romannumeral4} & LSTM & 19.87\\
		Experiment & Optimized LSTM & 9.13\\
		\hline
	\end{tabular}
\end{table}

\section{Conclusions}\label{conclusion}
\indent Based on the analysis of the characteristics of industrial big data, this paper improves the existing time series prediction algorithm, proposes the LSTM based on multivariate tuning, considering the characteristics of industrial data, especially its periodicity and multiple dimensions. It explains this new algorithm from multiple angles, including the structure and module design of the algorithm, pseudo code and the tuning process of the algorithm. Experiments with continuous data prediction algorithms show that the algorithm can effectively improve the accuracy of prediction.

\paragraph*{Acknowledgement}This paper was partially supported by NSFC grant U1509216, U1866602,61602129 and Microsoft Research Asia.

\bibliographystyle{ieeetr}
\bibliography{biblio}

\end{document}